\documentclass[letterpaper]{article} % DO NOT CHANGE THIS
\usepackage{aaai23}  % DO NOT CHANGE THIS
\usepackage{times}  % DO NOT CHANGE THIS
\usepackage{helvet}  % DO NOT CHANGE THIS
\usepackage{courier}  % DO NOT CHANGE THIS
\usepackage[hyphens]{url}  % DO NOT CHANGE THIS
\usepackage{graphicx} % DO NOT CHANGE THIS
\usepackage{natbib}  % DO NOT CHANGE THIS AND DO NOT ADD ANY OPTIONS TO IT
\usepackage{caption} % DO NOT CHANGE THIS AND DO NOT ADD ANY OPTIONS TO IT
\usepackage{algorithm}
\usepackage{algorithmic}
\usepackage{amsmath,amssymb,amsthm}
\usepackage{physics}
\usepackage{xcolor}
\usepackage{subfigure}
\newtheorem{definition}{Definition}

\newcommand{\akshat}[1]{{\color{black}#1}}

\renewcommand{\>}{\rangle}

\newcommand{\mc}{\mathcal}

\usepackage{newfloat}
\usepackage{listings}
\DeclareCaptionStyle{ruled}{labelfont=normalfont,labelsep=colon,strut=off} % DO NOT CHANGE THIS
\floatstyle{ruled}
\newfloat{listing}{tb}{lst}{}
\floatname{listing}{Listing}
\newcommand{\txn}{\mathcal{T}}

\title{Safe MDP Planning by Learning Temporal Patterns of Undesirable Trajectories and Averting Negative Side Effects}
\author{
    Siow Meng Low\textsuperscript{\rm 1}, 
    Akshat Kumar\textsuperscript{\rm 1}, 
    Scott Sanner\textsuperscript{\rm 2}
}
\affiliations{
    \textsuperscript{\rm 1} Singapore Management University\\
    \textsuperscript{\rm 2} University of Toronto\\
    smlow.2020@phdcs.smu.edu.sg, 
    akshatkumar@smu.edu.sg, 
    ssanner@mie.utoronto.ca
}

\usepackage{bibentry}
\usepackage{tabularx}

\begin{document}

\maketitle

\begin{abstract}
%Recent literature on Safe Markov Decision Process (MDP) Planning often focuses on safety requirements which are specifiable using the observed state and action representations. In actuality, the agent's state representation may lack sufficient fidelity to learn the safety aspects which can potentially be non-Markovian. In such scenario, unsafe sequence of actions can have non-immediate downstream impact on agent's safety and is not easily specifiable using only current state and action representation. In this paper, we revisit the constrained MDP formulation to accommodate such safety requirements. In particular, we employ a supervised learning model to learn the non-Markovian safety patterns which are challenging to be manually specified. We propose a model-based Lagrange Multiplier method, which incorporates the gradient knowledge learned in this supervised learning model, to facilitate agent learning of safe behaviors. Our empirical results show that this approach can satisfy complex non-Markovian safety constraints while optimizing agent's total returns. Our contributions are threefold: (i) deployment of a Supervised Learning Model which learns complex safe behaviors from full or partial trajectories, (ii) agent ability to exploit the gradient information learned in the supervised learning model, (iii) combination of Lagrange Multiplier method and Model-Based approach empowers agent to learn safe behaviors efficiently.
In safe MDP planning, a cost function based on the current state and action is often used to specify safety aspects. In the real world, often the state representation used may lack sufficient fidelity to specify such safety constraints. Operating based on an incomplete model can often produce unintended negative side effects (NSEs). To address these challenges, first, we associate safety signals with state-action trajectories (rather than just an immediate state-action). This makes our safety model highly general. We also assume categorical safety labels are given for different trajectories, rather than a numerical cost function, which is harder to specify by the problem designer. We then employ a supervised learning model to learn such non-Markovian safety patterns. Second, we develop a Lagrange multiplier method, which incorporates the safety model and the underlying MDP model in a single computation graph to facilitate agent learning of safe behaviors. Finally, our empirical results on a variety of discrete and continuous domains show that this approach can satisfy complex non-Markovian safety constraints while optimizing an agent's total returns, is highly scalable, and is also better than the previous best approach for Markovian NSEs.
\end{abstract}

\section*{Introduction}
\label{sec:intro}

In several environments, the agent must avoid both potential hazards while simultaneously optimizing its total accumulated reward. The existing literature in safe planning uses a constrained MDP formulation~\cite{altman2021constrained} and represents safety specifications as safety constraints that are derived from the immediately observable state and/or action~\cite{achiam2017constrained, tessler2018reward, stooke2020responsive, chow2019lyapunov, dalal2018safe,Simao2021alwayssafe}. However, obtaining a perfect description of the target environment becomes practically infeasible as autonomous agents are increasingly deployed in the real world~\cite{dietterich2017steps}. As a result, operating on such incomplete models may produce undesirable side effects, also called \textit{negative side effects} (NSEs), which are often discovered after agent deployment~\cite{amodei2016concrete,alizadeh2022considerate,krakovna2019penalizing,SKZijcai20,saisubramanian2022avoiding}. Therefore, addressing such NSEs has become a key challenge to increase the safety of deployed AI agents~\cite{saisubramanian2022avoiding}.

%Deep Reinforcement Learning (DRL) methods have garnered attention in the research community and have enjoyed great success in areas such as computer vision related task~\cite{Mnih2013PlayingAW} and continuous control planning~\cite{Low2022SampleefficientIL, wu2020scalable}. While DRL methods yield high quality policy, ensuring the eventual policy does not violate safety requirements remains a difficult problem. 
%The existing literature in safe RL largely represents safety specifications as safety constraints which is derived from the immediately observable state and/or action~\cite{achiam2017constrained, tessler2018reward, stooke2020responsive, chow2019lyapunov, dalal2018safe}. This formulation is not appropriate for settings where the agent's state representation lacks sufficient fidelity to learn the undesirable effects incurred by a policy~\cite{saisubramanian2022avoiding}, making the undesirable effects non-Markovian. For this purpose, we propose a new formulation where safety penalty is only observable by inspecting the entire history of an agent's trajectory. A safe policy is required to exhibit safe behaviors more often than a frequency threshold. 

One popular method used to solve MDPs with safety constraints is the Lagrange multiplier method~\cite{tessler2018reward, stooke2020responsive}. In this approach, the Lagrange multiplier (which acts as a penalty for constraint violation) is adapted slowly as the learning proceeds, converging to constraint-satisfying policy while also optimizing the primary reward objective. Other approaches include Lyapunov functions~\cite{chow2019lyapunov}, Trust Region methods~\cite{achiam2017constrained}, and constraint safety layer~\cite{dalal2018safe} methods. 
%Hyperparameter tuning for these approaches is particularly challenging compared to gradient-based Lagrangian method and some exhibit mixed performance gains over Lagrangian method. 
Notably, all these methods model safety requirements as functions of safety cost functions which are assumed to be known and  Markovian and hence functions of immediate state and action. Our work focuses on settings without such modelling assumptions, where we learn a classifier trained on trajectories labeled with different categories of safety labels for NSEs, hence modeling non-Markovian safety side effects.
%based on data associating state-action trajectories with NSEs, 
%that can classify any given trajectory into different categories of safely levels, 
%and is thus able to model non-Markovian safety effects. 
We then integrate this classifier with our safe planning approach. 

A closely related line of work for addressing NSEs is presented in~\cite{SKZijcai20, saisubramanian2022avoiding, shah2019preferences}.~\citet{saisubramanian2022avoiding} define NSE as \textit{undesired, unmodeled effects} due to incomplete MDP model specifications. Since not all undesirable effects can be foretold in advance, the model specification may lack sufficient fidelity to represent different types of NSEs. Their work proposed a supervised learning model to learn about NSEs through various types of human feedback data about NSEs, including human demonstration. Other research works propose different ways to infer \akshat{NSEs}, for instance through initial state configuration~\cite{shah2019preferences}, reachability of other states~\cite{krakovna2018penalizing} or attainable utility~\cite{turner2020conservative} after performing an action, ability to perform future task different from the current task~\cite{krakovna2020avoiding}. Bayes reasoning was also used in~\cite{hadfield2017inverse} to infer the true reward specification from a number of candidate reward functions. One common theme behind all these works is that not all NSEs can be anticipated precisely at design time and they need to be dynamically learned through human feedback or inferred. \akshat{Furthermore, all these works focus on \akshat{Markovian} NSEs, rather than a more general model where NSEs are associated with trajectories. It is a critical gap in previous approaches, such as~\cite{SKZijcai20}, as they decompose the penalty associated with an NSE into additive penalties associated with each state-action pair, which may not always be the case in a real world setting.} Alternatively, one can expand the state space definition to make NSEs Markovian, however this may make the agent's primary task computationally challenging due to complex state space, and knowledge about different types of NSEs is also not known apriori.

Another line of work~\cite{junges9636safety, alshiekh2018safe, jansen2020safe, jothimurugan2021compositional} has explored using logic specifications, such as temporal logic, to specify safety constraints which can be non-Markovian. In these methods, safety criteria need to be \textit{pre-specified}. As NSEs are side effects \textit{unmodeled} at the design stage, it is not feasible to pre-specify the NSE criteria and these methods are not suitable for NSE  setting. In contrast, our method only requires trajectory labels for NSE trajectories. 
%which are then fed to a general sequential learner to obtain a generalized representation. 
%If new NSE class (and criterion) is to be added, continual learning technique can be adopted without changing the MDP. 
In addition, temporal logic methods typically assume a safety-relevant abstraction of the original MDP (i.e. safety-relevant MDP)~\cite{alshiekh2018safe, jansen2020safe} which requires significant efforts from domain experts to construct, as opposed to just using the NSE labels. Lastly, our method learns to optimize reward and satisfy constraint jointly, whereas temporal logic methods synthesize a safety shield in-silos~\cite{alshiekh2018safe, jansen2020safe}. The shield either pre-specifies allowable actions or post-corrects the actions selected by the agent. This may lead to a sub-optimal strategy with reduced action space.

%Although NSE is often positioned as non-safety critical annoyances, we argue that certain safety requirements cannot be anticipated at design time either. This creates a need for safety constraints to be dynamically refined after deployment, inspiring us to propose a time series classification model to learn safety patterns from human experts.

%% 
%The approach proposed in this paper is inherently model-based since we make use of learned knowledge within a time series classification model for agent learning. Various model-based RL methods have been proposed in the past and one technique is to use a learned transition model to perform model unrolling. This will in turn yield synthetic samples to obtain a more accurate estimate of the policy objective~\cite{clavera2019model, amos2021model} or state-value function~\cite{feinberg2018model}, producing a better gradient estimate for the actor update. An inherent issue with this approach is that model error compounds over longer unrolling period~\cite{janner2019trust}, making it applicable to only short model rollouts and leaving the learned model pretty much unharnessed. Our paper adopts similar approach as Stochastic Value Gradients~\cite{heess2015learning} where we collect real trajectory samples and exploiting model gradients evaluated at these real data points. 

Our main contributions are the following. \textit{First}, we formulate safe MDP planning problem as constrained MDP~\cite{altman2021constrained}. Unlike previous methods, we do not assume a Markovian safety cost function is given. We utilize a supervised learning model to learn safety characteristics of a trajectory from the NSE data. \textit{Second}, we integrate the learned safety model and the MDP model in a single computation graph, and develop a Lagrange multiplier~\cite{Bertsekas99} method to optimize the policy while respecting trajectory-based safety constraints. We (a) proposed a method applicable to both Markovian and non-Markovian NSEs, and discrete and continuous domains, with much higher scalability than the previous method~\cite{SKZijcai20}; (b) developed a model-free approach and show that our model-based method is significantly better. \textit{Finally}, our empirical results on a variety of discrete and continuous domains show that our highly scalable approach can satisfy complex non-Markovian safety constraints, while optimizing agent's total returns, and outperform previous best approach for Markovian NSEs.

%where safety constraints cannot be precisely quantified beforehand and thus utilized a supervised learning model to learn safety characteristics of a trajectory from human expert. Secondly, our method collects real sample trajectories and models are only used to gradient backpropagation in learning safe behaviors. Lastly, we propose a novel method to update the Lagrange Multiplier in a sample-efficient manner.

\section*{Problem Formulation}
\label{sec:problem}

Sequential decision making under uncertainty is typically represented using Markov Decision Processes (MDPs)~\cite{sutton1998introduction}. Constrained Markov Decision Processes (CMDPs)~\cite{altman1998constrained} allow incorporation of constraints in the problem formulation, which for example can model safety aspects in decision making. A CMDP can be defined using tuple $(S, A, \txn, R, \gamma, b_0, \mc{C}, \mc{D})$. We assume a general setting with continuous state and action spaces ($S\subseteq \mathbb{R}^n$, $A\subseteq \mathbb{R}^m$). The environment state transition is characterized by the function $p(s_{t+1}| s_t, a_t) \!=\!\txn(s_t, a_t, s_{t+1})$. A reward function $R: S \times A \rightarrow \mathbb{R}$ maps a state, action to a scalar reward value; $\gamma\in [0, 1)$ is the reward discount factor, and $b_0$ is the initial state distribution. We assume a model-based setting where different model components are known (e.g., reward, transition function).

Constraints are represented using cost functions $c_{i}\in \mc{C}$, and cost limits $d_{i}\in \mc{D}$ on the total expected costs. Each  function $c_i(s, a)$ maps a state-action tuple to a real valued cost. Often, cost functions and constraints are used to represent safety aspects in a problem domain.

\vskip 2pt
\noindent{\textbf{Policy: }}We consider a probabilistic policy parameterized by a rich function approximator, such as a deep neural network. A probabilistic parameterized policy is denoted as $\pi_{\theta}(s, a)$ which refers to the probability of executing action $a$ in state $s$; $\theta$ refers to the parameters (to-be-optimized) of the policy function approximator.

%Recent safe RL literature~\cite{tessler2018reward, stooke2020responsive} requires these cost functions to be function of three variable sets: state, action, next state. Instead, this paper defines cost function as function of policy parameters $\theta$, i.e. $c_{i}(\theta)$.

\vskip 2pt
\noindent{\textbf{The CMDP problem: }}The \textit{primary objective} of a CMDP is to maximize the expected sum of discounted rewards over trajectories $\tau = (s_0, a_0, s_1, a_1, ...)$ by following a parameterized policy $\pi_{\theta}$. The objective is~\cite{achiam2017constrained}: $$J_{R}(\pi_{\theta}) = \mathbb{E}_{\tau \sim \pi_{\theta}} [ \sum_{t=0}^{\infty} \gamma^{t} R(s_t, a_t) ]$$ where $\tau\sim \pi_{\theta}$ indicates that trajectories are sampled from the distribution induced by the policy  $\pi_{\theta}$, initial state distribution $b_0$, and the state transition function.

Similar to $J_{R}$, we can define the expected value of a cost function $c_i\in \mc{C}$ over the induced trajectories:
\begin{align}
    J_{c_i}(\pi_{\theta}) = \mathbb{E}_{\tau \sim \pi_{\theta}} [\sum_{t=0}^{\infty} \gamma^{t} c_i(s_t, a_t)] \label{eq:cost}
\end{align}
%$J_{C}(\pi_{\theta}) = \mathbb{E}_{\tau \sim \pi_{\theta}} [c(\tau)]$. 
The CMDP problem is expressed as: 
\begin{align}
	&\max_{\theta}  J_{R}(\pi_{\theta}) \\
	&\text{s.t.} \quad J_{c_i}(\pi_{\theta}) \leq {d_i} \quad \forall c_i\in \mc{C}
\end{align}
where $d_i\in \mc{D}$ is the corresponding threshold for $J_{c_i}$.

 %We make the same set of assumptions as~\cite{tessler2018reward} to ensure convergence to feasible solution. 

There are some key implicit assumptions in such a CMDP formulation. First, it is assumed that the cost functions $c_i$, which for example can define the safety aspects, are known. This may not always be the case in a real world setting as obtaining the exact numerical specification of safety cost functions is challenging. Second, it is assumed that the expected safety cost in~\eqref{eq:cost} is decomposed and additive per time step. As noted in section~\ref{sec:intro}, this may not be always feasible if the model specification is incomplete, which may make safety as a function of the entire trajectory $\tau$, rather than just the immediate state and action. To address these issues, next we discuss a formulation in which data about safety aspects is collected in the form of negative side effects (NSEs), and a safety model is learned from such a dataset.

\subsection*{Negative Side Effects (NSEs)}
\label{sec:nse}

Our NSE definition is similar to~\citet{saisubramanian2022avoiding} where we consider episodic tasks and each complete trajectory $\tau\!=\!\<s_0, a_0, s_1, \ldots, a_{T}, s_{T}\>$ can be of arbitrary (but finite) length, terminating in state $s_T$.
\begin{definition}
\label{def:partition}
Let $\Psi$ denote a \emph{partition} of the space of all possible complete trajectories for the given MDP. The set $\Psi$ can be further divided into $K$ mutually exclusive sets of NSE \emph{categories}: $\Psi = \psi_1\cup \psi_2\cup\ldots \cup \psi_K$.
\end{definition}
Each $\psi_i$ defines a category of NSEs (e.g., mild, moderate, severe, or no-NSE). Without loss of generality, we assume that a trajectory only belongs to a \emph{single category} of NSEs (e.g., the most severe form of NSE occurring in the trajectory). When a trajectory $\tau\in \psi_i$ is observed during execution, NSE $i$ is said to have occurred. We consider \textit{non-Markovian} NSEs as NSE severity can depend on the entire trajectory, in contrast to Markovian NSEs which depend on a single state-action pair~\cite{SKZijcai20}. Non-Markovian NSEs are more general than Markovian NSEs, and can model complex NSEs without the need to expand the state representation of the MDP. Expanding the state representation to ensure Markovian NSEs may introduce new risks and require extensive data collection/evaluation, which is undesirable.  

We also note several modeling benefits of our method over prior works~\cite{SKZijcai20,saisubramanian2022avoiding}. Although prior works introduce non-Markovian NSEs, they do not present an approach to solve it. We use powerful neural network based function approximators that can scale well to large state spaces, unlike tabular policies used in prior works, which are not scalable to large state spaces. Previous work is also limited to discrete state and action spaces, whereas our method can handle both discrete and continuous problems. In prior works, one needs to chose an arbitrary numerical penalty score for different types of NSEs, and it is unclear how to choose it, whereas in our work, we directly exploit different categories of NSEs (e.g., mild, moderate, no-NSE), which is relatively easy to specify (e.g., by elicitation from domain experts) in comparison.

%Expanding the state representation to ensure Markovian NSE may introduce new risks and require extensive data collection/evaluation, which is undesirable.  

% MDP Transition and Reward is Markovian
% Non-Markovian in nature

\subsection*{Trajectory Classification with RNNs}
\label{sec:rnn}

In general, it is infeasible to exactly define each NSE partition $\psi_i$ beforehand. As noted in~\cite{SKZijcai20,saisubramanian2022avoiding}, as agents are deployed in the real world, data about different NSEs can be collected. 
%In addition, several other strategies to collect data about different NSEs, including random exploration, human feedback among others, are presented in~\cite{SKZijcai20}. 
We assume that such a dataset is available that contains trajectories for different NSE categories $k=1:K$ (as noted in definition~\ref{def:partition}). A data entry in this set, $(\tau, k)$, implies NSE category $k$ is associated with the state-action trajectory $\tau$. %Empirically, 
In our empirical experiments, we optimized a policy without %considering constraints 
averting NSEs for a domain, and %using this policy collected data about NSEs. 
subsequently used the trajectory data during the learning process for NSE labeling. This approach resembles incremental learning process in practice, where an agent interacts with the environment without first being aware of the NSE. The historical trajectory data is collected and labeled with the corresponding NSE class (including no-NSE). The labeled data entries $(\tau, k)$ can then be used for classifier training.

\begin{figure}[t]
	\centering
	\includegraphics[scale=0.59]{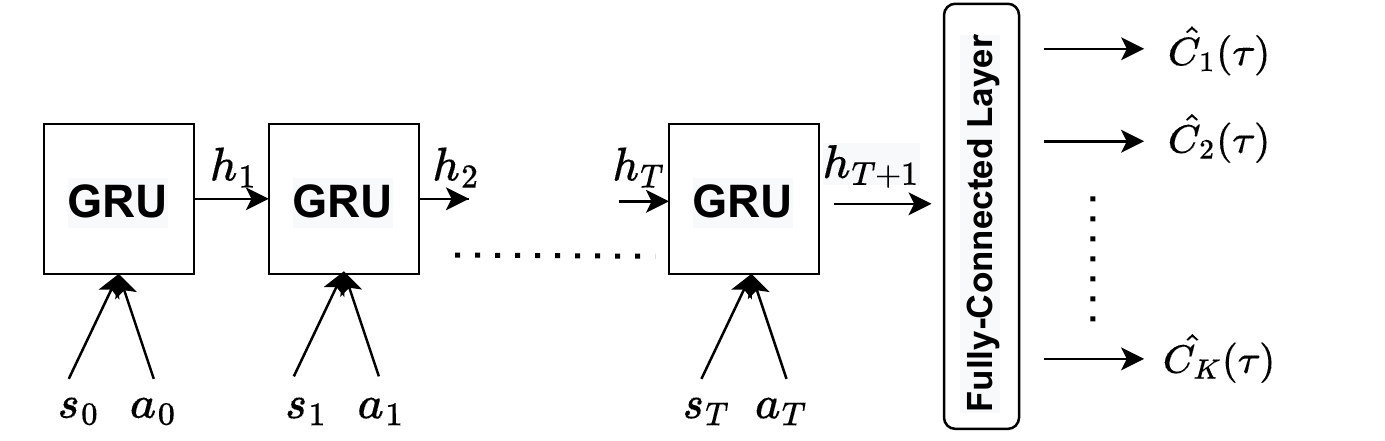}
	\vskip 0pt
	\caption{NSE Classifier Using Recurrent GRUs}
	\label{fig:gru}
	\vskip 0pt
\end{figure}

As the trajectory $\tau$ can be of variable length for episodic tasks, we propose an RNN - Recurrent Neural Network~\cite{rumelhart1986learning} in classifying a trajectory $\tau$ into its respective NSE category $k$. 
We used Gated Recurrent Units (GRU) as our specific RNN architecture, which balances between accuracy and network parameter size~\cite{cho2014properties}. The GRU network used in our empirical experiment is depicted in Figure~\ref{fig:gru}.
%\akshat{As different trajectories $\tau$ can be of different length, we use a recurrent neural network~\cite{XXX} to classify a trajectory into a NSE category.include RNN figure showing how state and action for a $\tau$ are given as input, and what is output. TOtal there are $K$ NSE categories}

%% 
We also note that the trajectory classification method we have presented above allows a modular separation between optimizing the primary task of the underlying MDP, and safety specifications. As more NSE data is gathered, we can update our NSE classifier with additional data, without the need to always change the underlying state space and reward function each time new NSEs are discovered. This enables modular and continual safety learning, which is critical for real world deployment of autonomous agents.

% SCG, two derivatives, trajectory classifier, NSE

\subsection*{CMDP for Avoiding Non-Markovian NSEs}

As noted in section~\ref{sec:problem}, the CMDP objective is to maximize the expected sum of discounted rewards over trajectories $\tau$. The additional NSE constraints imposed are to limit the occurrence probability of trajectories with undesirable NSEs. Note that we consider NSEs which are non-Markovian and depend on the entire trajectory. Thus, the non-Markovian NSE based CMDP problem can be expressed as:
\begin{equation}
\begin{aligned}
 \max_{\theta} & \: J_{R}(\pi_{\theta}) = \mathbb{E}_{\tau \sim \pi_{\theta}} [ \sum_{t=0}^{\infty} \gamma^{t} R(s_t, a_t) ] \\
 \textrm{s.t.} & \: J_{c_i}(\pi_{\theta}) = \mathbb{E}_{\tau \sim \pi_{\theta}} [ C_{i}(\tau) ] \leq d_i \quad \forall i \in {1, 2, ..., K} \label{eq:cmdp}
\end{aligned}
\end{equation}
In this formulation, $K$ refers to the number of NSE categories while $0 \leq {d}_{i} \leq 1$ specifies the tolerable proportion of trajectories having NSE of class $i$. The value ${d}_{i}$ is  application-dependent and typically configured by the practitioner. $C_{i}(\tau)$ is an indicator function to denote the presence of NSE of class $i$ in the given trajectory $\tau$. As NSEs are unmodeled side effects, they may not have a closed-form expression. Instead, it needs to be estimated using labelled trajectory data as noted in section~\ref{sec:rnn}. In this paper, we approximate the ground-truth $C_{i}(\tau)$ using RNN-based trajectory classifier output $\hat{C}_{i}(\tau)$, which can be interpreted as the estimated probability of NSE of class $i$ occurring in trajectory $\tau$. The approximate CMDP problem using trajectory classifier can be rewritten as follows:
\begin{equation} \label{cmdp}
\begin{aligned}
 \max_{\theta} & \: J_{R}(\pi_{\theta}) = \mathbb{E}_{\tau \sim \pi_{\theta}} [ \sum_{t=0}^{\infty} \gamma^{t} R(s_t, a_t) ] \\
 \textrm{s.t.} & \: J_{\hat{c}_i}(\pi_{\theta}) = \mathbb{E}_{\tau \sim \pi_{\theta}} [ \hat{C}_{i}(\tau) ] \leq d_{i} \quad \forall i \in {1, 2, ..., K} 
\end{aligned}
\end{equation}

\section*{Proposed Method}

We next present a Lagrangian relaxation based method~\cite{Bertsekas99}, which has been popular recently to solve CMDPs. We will primarily focus on new techniques that enable us to handle trajectory based constraints $J_{\hat{c}_i}(\pi_{\theta})\leq d_i$ in~\eqref{cmdp}, which are complex to optimize over as they involve a RNN-based trajectory classifier (i.e., $\hat{C}_{i}$), and are non-additive and non-decomposable among time steps, unlike the case in standard CMDPs (see Eq.~\eqref{eq:cost}).

\subsection*{Lagrangian Method}

We apply the Lagrange multiplier method to solve the CMDP problem~\eqref{cmdp}. The Lagrange multiplier method is shown to converge to the local optimum of CMDP problems and performs well empirically~\cite{tessler2018reward,stooke2020responsive}. The  Lagrangian dual problem of~\eqref{cmdp} is written as: 
\begin{align} \label{Lagrangian}
 \min_{\boldsymbol{\lambda} \geq 0} \max_{\theta} & \: L(\boldsymbol{\lambda}, \theta) = J_{R}(\pi_{\theta}) - {\sum}_{i=1}^{K} {\lambda}_{i} [J_{\hat{c}_i}(\pi_{\theta}) - {d_i}]
\end{align}
We can also compute the gradient of Lagrangian function $L$ w.r.t. different Lagrange multipliers $\lambda_i$ and the policy parameters $\theta$ as:
\begin{align} 
 -\frac{\partial{L}}{\partial{{\lambda}_{i}}} & \: = J_{\hat{c}_i}(\pi_{\theta}) - {d_i} = \mathbb{E}_{\tau \sim \pi_{\theta}} [ \hat{C}_{i}(\tau) - {d}_{i} ] \label{eq:mul}
\end{align}

\begin{equation} \label{theta_grad}
\begin{aligned}
 \frac{\partial{L}}{\partial{\theta}} & \: = \nabla_{\theta} J_{R}(\pi_{\theta}) - {\sum}_{i=1}^{K} {\lambda}_{i} \nabla_{\theta} J_{\hat{c}_i}(\pi_{\theta}) \\
 & \: = \nabla_{\theta} J_{R}(\pi_{\theta}) - {\sum}_{i=1}^{K} {\lambda}_{i} \nabla_{\theta} \mathbb{E}_{\tau \sim \pi_{\theta}} [ \hat{C}_{i}(\tau) ]
\end{aligned}
\end{equation}

% We can also compute the gradient of Lagrangian w.r.t. different Lagrange Multipliers $\nu_i$ and the policy parameters $\theta$ as:
% \begin{align}
% -\frac{\partial{L}}{\partial{{\nu}_{i}}} & \: = J_{\hat{c}_i}(\pi_{\theta}) - {d_i} = \mathbb{E}_{\tau \sim \pi_{\theta}} [ \hat{C}_{i}(\tau) - {d}_{i} ] \\
%  = \nabla_{\theta} J_{R}(\pi_{\theta}) - {\sum}_{i=1}^{K} {\nu}_{i} \nabla_{\theta} \mathbb{E}_{\tau \sim \pi_{\theta}} [ \hat{C}_{i}(\tau) ]
% \end{align}

%The unconstrained optimization problem \(L\) in~\eqref{Lagrangian} is called the Lagrangian function while equations~\eqref{nu_grad} and~\eqref{theta_grad} are the gradient of the Lagrangian wrt the multipliers and policy parameters respectively.
Optimizing the  Lagrangian problem~\eqref{Lagrangian} can provide a local optimum of the problem~\eqref{cmdp}.
The vector, \(\boldsymbol{\lambda} = [{\lambda}_1 \ {\lambda}_2 \ ... \ {\lambda}_K] \geq \boldsymbol{0}\), contains the Lagrange multipliers for each of the \(K\) constraints. Each Lagrange multiplier serves as the tradeoff coefficient for the respective constraint. 
 %When the Lagrangian is minimized wrt the multipliers, the solution attained will be optimal and satisfy the constraints. 
Learning to solve this Lagrangian problem, as shown in~\cite{tessler2018reward}, typically involves updating \(\boldsymbol{\lambda}\) and \(\theta\) using the gradient expressions derived in~\eqref{eq:mul} and~\eqref{theta_grad} (we refer the reader to~\citet{tessler2018reward} for details). 

There are a number of methods estimating the first gradient term \(\nabla_{\theta} J_{R}(\pi_{\theta})\) in~\eqref{theta_grad}, including model-based Stochastic Value Gradients~\cite{heess2015learning}, iterative lower bound optimization~\cite{Low2022SampleefficientIL}, and model-free Proximal Policy Optimization (PPO)~\cite{DBLP:journals/corr/SchulmanWDRK17}. The empirical study in this paper has been conducted using PPO as it provided better performance compared to others in terms of achieving good solution quality with higher sample efficiency. Next, we mainly focus on estimating the second gradient term \(\nabla_{\theta} \mathbb{E}_{\tau \sim \pi_{\theta}} [ \hat{C}_{i}(\tau) ]\). This is much more challenging than the standard Lagrangian method for CMDPs~\cite{tessler2018reward} as $\hat{C}_{i}$ is an RNN output, rather than an additive sum of costs as in a standard CMDP (see Eq.~\eqref{eq:cost}).

%Prior work defines the constraint term as an aggregate of Markovian penalties where \(\hat{C}_{i}(\tau)\) in~\eqref{Lagrangian} refers to the estimated probability of NSE present in a given trajectory \(\tau\). In the subsequent subsections, We analyze and discuss two different methods in estimating the gradients of these terms.

\subsection*{Model-Free Gradient Estimation} \label{section_mfge}

The second term $\nabla_{\theta} \mathbb{E}_{\tau \sim \pi_{\theta}} [ \hat{C}_{i}(\tau) ]$
%\({\sum}_{i=1}^{K} {\nu}_{i} \nabla_{\theta} \mathbb{E}_{\tau \sim \pi_{\theta}} [ \hat{C}_{i}(\tau) ]\)
of~\eqref{theta_grad} involves a gradient of an expectation taken over a random variable. The score function estimator~\cite{fu2006gradient} makes use of the log derivative trick and estimates the gradient using the output score from classifier. The score function estimator is derived as: 

%\begin{equation} 
\begin{align} \label{sf_estimator}
 & \nabla_{\theta} \mathbb{E}_{\tau \sim \pi_{\theta}} [ \hat{C}_{i}(\tau) ] \nonumber \\ 
 = & \nabla_{\theta} \Big [ \mathbb{E}_{\tau \sim \pi_{\theta}} [ \hat{C}_{i}(\tau) ] - {d}_{i} \Big ] \nonumber \\
 = & \nabla_{\theta} \mathbb{E}_{\tau \sim \pi_{\theta}} [ \hat{C}_{i}(\tau) - {d}_{i} ] \nonumber \\
 = & \mathbb{E}_{\tau \sim \pi_{\theta}} [ (\hat{C}_{i}(\tau) - {d}_{i}) \nabla_{\theta} \log P(\tau; \theta) ] \nonumber \\
 = & \mathbb{E}_{\tau \sim \pi_{\theta}} [ (\hat{C}_{i}(\tau) - {d}_{i}) \sum_{t=0}^{T} \nabla_{\theta} \log \pi_{\theta}(s_t, a_t) ]
\end{align}
%\end{equation}

\(T\) in~\eqref{sf_estimator} is the maximum timestep encountered in a particular trajectory \(\tau\). The \(-d_i\) term, introduced at the first step, serves as a baseline to reduce the variance of this estimator. This estimator provides a clever way to collect on-policy samples, estimate gradients wrt \(\theta\) and adjust the policy parameters $\theta$ accordingly.

Before discussing the interpretation of the estimator in~\eqref{sf_estimator}, we first highlight that the policy parameters are adjusted in the opposite direction of this gradient due to the negative sign in~\eqref{theta_grad}. \akshat{Consider one particular trajectory \(\tau\) from the minibatch samples collected in~\eqref{sf_estimator}. When \(\hat{C}_{i}(\tau) - {d}_{i}\) is positive, the policy update will move in the opposite direction of \(\sum_{t=0}^{T} \nabla_{\theta} \log \pi_{\theta}(s_t, a_t)\), decreasing the likelihood of performing similar sets of actions in future}. Conversely, when \(\hat{C}_{i}(\tau) - {d}_{i}\) is negative, the incremental update will increase the likelihood of performing such actions. 

\begin{figure*}[htb]
	\centering
	\includegraphics[width=0.91\textwidth]{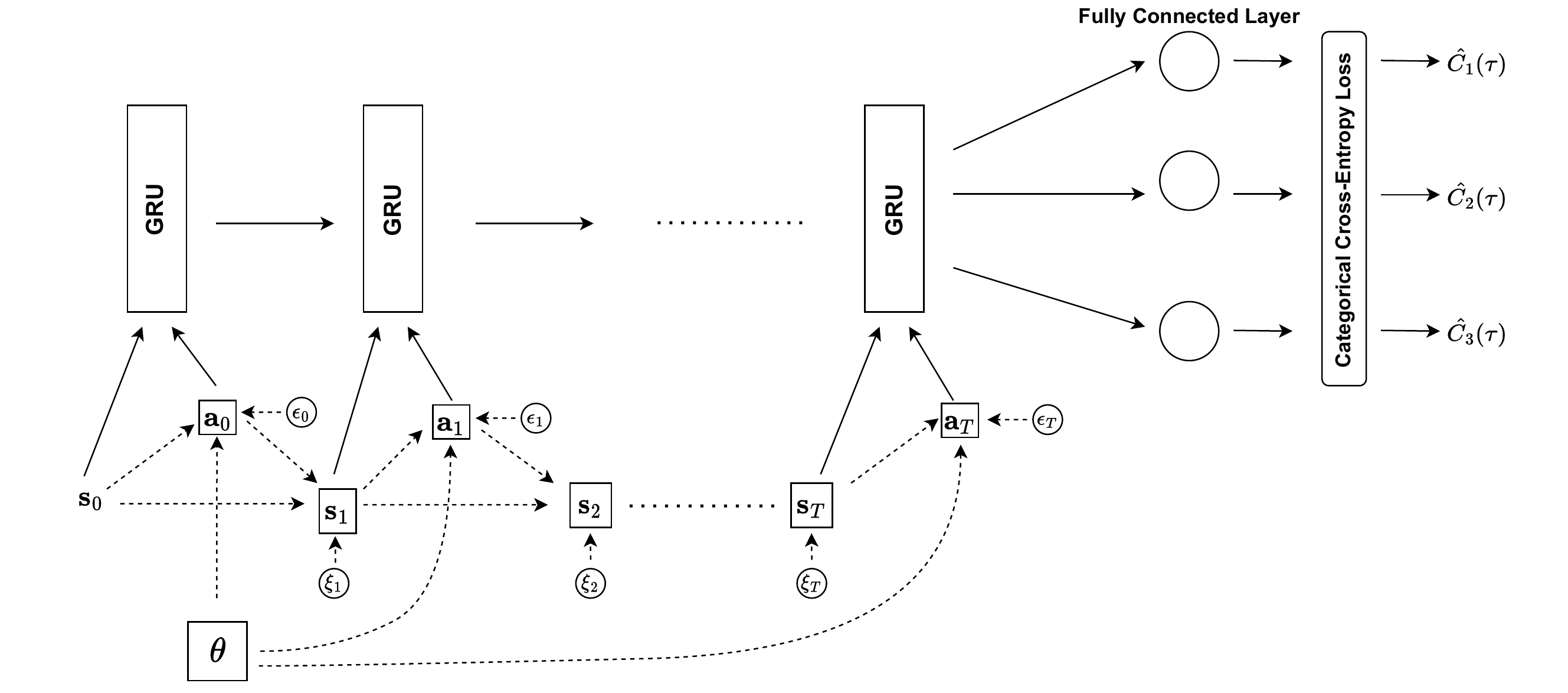}
	\caption{Stochastic Computation Graph of \(\hat{C}_{i}(\tau)\) for trajectory \(\tau\) where number of NSE classes is \(K = 3\)}
	\label{fig:SCG}
\end{figure*}

Although the model-free method provides a neat way of estimating the gradient, it is rather difficult to apply in practice due to the following reasons. Firstly, score function estimators tend to have high variance~\cite{10.5555/2969442.2969633}, in turn increasing the difficulty of moving towards the feasible constraint region in a consistent manner. Secondly, the output \(\hat{C}_{i}(\tau)\) is an approximate value of the unknown true value \({C}_{i}(\tau)\). The classification error exacerbates the high variance issue. \akshat{Lastly, the score function estimator sums up the gradients evaluated at all individual timesteps, i.e. \(\sum_{t=0}^{T} \nabla_{\theta} \log \pi_{\theta}(s_t, a_t)\). This implies that the policy will be adjusted to decrease or increase (depending on the sign of \(\hat{C}_{i}(\tau) - {d}_{i}\)) the likelihood of performing the sampled actions at every single timestep.} Even though the presence of an NSE is determined by inspecting the entire trajectory \(\tau\), the contributing factor could be a smaller subpart of the trajectory. In such cases, adjusting the actions at every single timestep can lead to over-correction. %and make it harder to reach the specified threshold \(d_i\). 

In light of the difficulties in applying model-free gradient estimation methods, we now propose a model-based gradient estimation method which better exploits the differentiable function approximator learned in \(\hat{C}_{i}(\tau)\).

\subsection*{Model-Based Gradient Estimation} \label{section_mbge}

The pathwise derivative (PD) estimator~\cite{glasserman2004monte,10.5555/2969442.2969633} has lower variance but it requires (a) \(\hat{C}_{i}(\tau)\) be a continuous function of \(\theta\), (b) \(\tau\) be a deterministic and differentiable function of \(\theta\). To fulfil these conditions, we apply the ``reparameterization trick'' on both the transition model \(\txn(s_t, a_t, s_{t+1})\) and the policy $\pi_{\theta}(s, a)$. 

Figure \ref{fig:SCG} illustrates such a reparameterized model using a stochastic computation graph~\cite{10.5555/2969442.2969633}. Figure's upper half depicts the GRU network used in classifying trajectory $\tau$, which outputs $\hat{C}_{i}(\tau)$. The bottom half outlines the relationship between policy parameters $\theta$ and the state-action pairs ($s_t, a_t$) observed in trajectory $\tau$. To express $s_t, a_t$ as deterministic and differentiable function of $\theta$, the reparameterization trick uses exogenous random samples $\xi, \epsilon$ to sample from the probability distributions $\txn(s_t, a_t, s_{t+1})$ and $\pi_{\theta}(s, a)$ respectively. For location-scale probability distributions (e.g. Gaussian, Gamma), sampled state can be written~\cite{KingmaW13} as $s_{t+1}(\theta, \xi) = \mu(s_t, a_t; \theta) + \sigma(s_t, a_t; \theta) \cdot \xi_{t+1}$ where $\mu$, $\sigma$ refer to the location and scale respectively while $\xi_t$ is a random sample drawn from a normalized distribution with zero location and unit scale. For categorical distributions, Gumbel softmax~\cite{jang2016categorical} can be employed to rewrite sampled action $a_{t}(\theta, \epsilon) = {one\_hot} (\text{softmax} (\log {\pi_{\theta}(s_t, a)} + \epsilon_t))$ with $\epsilon_t$ drawn from Gumbel(0, 1). Sampled state can be rewritten similarly.

We call this method model-based gradient estimation as the transition model $\txn$ is required for the reparameterization trick. The pathwise derivative estimator is written as: 
%\begin{equation} 
\begin{align}
 & \nabla_{\theta} \mathbb{E}_{\tau \sim \pi_{\theta}} [ \hat{C}_{i}(\tau) ] \nonumber \\ 
 = & \nabla_{\theta} \mathbb{E}_{\tau \sim \pi_{\theta}} [ \hat{C}_{i}(s_0, a_0, s_1, a_1, ..., s_T, a_T) ] \nonumber \\
 = & \nabla_{\theta} \mathbb{E}_{\boldsymbol{\epsilon},\boldsymbol{\xi} \sim p(\cdot)} [ \hat{C}_{i}(s_0, \theta, \boldsymbol{\epsilon}, \boldsymbol{\xi}) ] \label{pd_reparam} \\
 = & \mathbb{E}_{\boldsymbol{\epsilon},\boldsymbol{\xi} \sim p(\cdot)} [ \nabla_{\theta} \hat{C}_{i}(s_0, \theta, \boldsymbol{\epsilon}, \boldsymbol{\xi}) ] \label{pd_estimator}
\end{align}
%\end{equation}

\begin{table*}[th]
\small 
\begin{center}
\begin{tabular}{  c | c | c | c | c  }
\hline
% \multicolumn{1}{|c|}{label01} & \multicolumn{3}{c|}{label02}
\multicolumn{1}{c|}{} & \multicolumn{2}{c|}{Boxpushing} & \multicolumn{2}{c}{Driving} \\
\hline
 & Reward & NSE & Reward & NSE \\
\hline
Markov-HA-S & $-30.43 \pm 2.29$ & $0.76 \pm 3.40$ & $-26.59 \pm 0.46$ & $8.24 \pm 8.75$ \\
MF-Lagrange & $-28.29 \pm 1.92$ & $59.00 \pm 10.95$ & $-28.10 \pm 0.98$ & $5.10 \pm 7.46$ \\
\textbf{MBGE} & $\mathbf{-26.80 \pm 0.89}$ & $\mathbf{0.30 \pm 0.72}$ & $\mathbf{-25.78 \pm 1.01}$ & $\mathbf{2.97 \pm 6.61}$  \\
\hline
\end{tabular}
\caption{Grid World Experiment Results - Boxpushing \& Driving}
\label{table:grid_results}
\end{center}
\end{table*}

The reparameterization trick is performed in step~\eqref{pd_reparam} to rewrite $\hat{C}_{i}(\tau)$ as a function of the initial state $s_0$, policy parameters $\theta$ and the exogenous noise vectors $\boldsymbol{\epsilon}, \boldsymbol{\xi}$ (of size $T + 1$ and size $T$ respectively) drawn from probability distribution independent of $\theta$. Recall that classifier output is a function of trajectory $\tau = (s_0, a_0, s_1, a_1, ..., s_T, a_T)$. All intermediate $s_k$ (except $s_0$) has a recursive relationship with the previous state-action pair ($s_{k-1}, a_{k-1})$ and can be rewritten as a function of $\theta$ and $\boldsymbol{\xi}$ using the reparamterization formulae described earlier. Similarly, $a_k$ is dependent on $s_k$ and $\theta$. Since $s_k$ can be written as a function of $\theta$ and $\boldsymbol{\xi}$, the observed $a_k$ (sampled from probability parameterized policy $\pi_{\theta}(s, a)$) can also be written as a function of $\theta, \boldsymbol{\epsilon}$ and $\boldsymbol{\xi}$ in a similar manner. As the function approximator $\hat{C}_{i}(\tau)$ provides a continuous and differentiable function, the model-based gradient estimation method exploits this and averages the sample derivative $\nabla_{\theta} \hat{C}_{i}(s_0, \theta, \boldsymbol{\epsilon}, \boldsymbol{\xi})$ in the minibatch. 

We also discuss the characteristics of this estimator. First, model-based gradient estimation utilizes the derivative $\nabla_{\theta} \hat{C}_{i}$ instead of the actual output $\hat{C}_{i}$. Using the derivative typically results in lower variance~\cite{10.5555/2969442.2969633,heess2015learning}, especially when the function $\hat{C}_{i}$ is smooth. Second, the proposed trajectory classifier uses the cross-entropy loss function, causing the gradient $\nabla_{\theta} \hat{C}_{i}$ to gradually diminish when predicted probability $\hat{C}_{i}$ approaches 1 or 0. The small gradient values encountered will impede the gradient update of the Lagrange Multiplier term. Regularization (e.g. Dropout) is recommended to prevent the classifier from outputting a value very close to 1 or 0 (e.g. $\geq 0.9999$). In addition, we recommend a non-zero initial value for $\boldsymbol{\lambda}$ to prevent $\theta$ from quickly converging to an infeasible region where the evaluated gradient $\nabla_{\theta} \hat{C}_{i}$ is very small. Last, for the computation of $\hat{C}_{i}$, the trajectory classifier can have different weights for the state-action pair $(s_t, a_t)$ at different timestep $t$. This permits the policy network to target specific timesteps which greatly impacts the final score $\hat{C}_{i}$ and adjust the policy parameters $\theta$ using the gradients evaluated at these timesteps accordingly. This is beneficial and tackles the over-correction issue discussed in model-free gradient estimation. 

%Gradient \(\nabla_{\theta} J(\theta)\) is estimated using gradient from critic \(\hat{Q}(s, {\mu}_{\theta}(s))\) while gradient \(\nabla_{\theta} C(\theta)\) is estimated using NSE classifier \(\hat{C}(\tau)\) and stochastic computation graph. NSE classifier provides \(\frac{\partial{\hat{C}}}{\partial{\textbf{s}}}\)and \(\frac{\partial{\hat{C}}}{\partial{\textbf{a}}}\). The stochastic computation graph chains these gradient back to actor network by computing \(\frac{\partial{\textbf{s}}}{\partial{\boldsymbol{\theta}}}\) and \(\frac{\partial{\textbf{a}}}{\partial{\boldsymbol{\theta}}}\).  
%with chain rule \(\frac{\partial{\hat{C}}}{\partial{\theta}} = \frac{\partial{\hat{C}}}{\partial{\textbf{a}}} \frac{\partial{\textbf{a}}}{\partial{\theta}}\).

\section*{Empirical Experiments}

To test the effectiveness of our proposed methods\footnote{Code available on: \url{https://github.com/siowmeng/Avert-NonMarkovNSE}}, we present two sets of experiment results. The first set of experiments involves two different tasks in a grid-world environment where state and action spaces are both discrete. The objective of this experiment is to compare against the current state-of-the-art safe planning approach~\cite{saisubramanian2022avoiding} which avoids Markovian NSEs. We show that our proposed model-based gradient estimation (MBGE) method produces better results in balancing between maximizing reward and averting NSEs. The second set of experiments evaluates the performance in performing two distinct tasks in continuous MDP planning domains. The detailed definitions of these continuous MDP planning domains can be found in~\cite{bueno2019deep,rddlgym2020}. Our proposed method successfully optimizes rewards while reducing the occurrence of non-Markovian NSEs below the specified thresholds. 

To perform gradient-based updates, our proposed method collects a batch of trajectory samples in a learning epoch and each experiment run trains the agent for a number of learning epochs (1000 for Grid-Worlds, 5000 for continuous domain). During training, after a fixed number of learning epochs, 100 separate test trajectories (different from the training trajectories) are collected for performance evaluation. \akshat{The rest of hyperparameter settings used in our experiments are provided in the supplement.} 

For performance reporting, we report the average and standard deviation values for 5 training runs. This is because our approach involves learning from samples collected from stochastic environments, while conventional planning approach like~\cite{saisubramanian2022avoiding} does not collect trajectory samples. To have a fair estimate of the planning performance of our approach, we evaluate the average performance over the 5 runs.

% \begin{figure}[t]
% 	\centering
% 	\includegraphics[scale=0.47]{Figures/grid_world.pdf}
% 	\vskip 0pt
% 	\caption{Grid World Experiments - Boxpushing \& Driving}
% 	\label{fig:grid_expt}
% 	\vskip -5pt
% \end{figure}

\subsection*{Grid World Domains}

Here we provide a brief description of the grid world domains used to test our approach against current state-of-the-art Markovian NSE planner which learns NSE from human-provided numerical score of a state-action pair. We call this method \textit{Markov-HA-S} and refer the readers to~\cite{saisubramanian2022avoiding} for more comprehensive description of the experiment setup. Experiments have been carried out for two separate tasks: 

\subsubsection{Boxpushing} The agent is required to pick up a box at a specific location and move towards a goal position~\cite{10.5555/3020488.3020530}. The agent will incur NSEs if it lands on certain surface types (e.g. surface with carpet, fragile surface) while pushing the box. 

\subsubsection{Autonomous Driving} The autonomous agent is incentivized to move toward a goal position as quickly as position~\cite{Wray_Zilberstein_Mouaddib_2015}. However, it will incur NSEs if it tries to drive fast while nearing a pedestrian or a puddle. 

\subsubsection{Negative Side Effects} For \textit{Markov-HA-S}, NSEs are defined as Markovian NSEs which are immediately observable for a given state-action pair. To compare against their result, we adapt our classifier architecture and make it a Multi-Layer Perceptron (MLP) classifier~\cite{GoodBengCour16} which classifies NSE for a given state-action pair, i.e. $\hat{C_i}(s, a)$. Similar to \textit{Markov-HA-S}, experiments were conducted in fully avoidable NSE scenarios and thus our constraint can be adapted as $\sum_{t=0}^T \hat{C_i}(s_t, a_t) \leq 0$. %We compare our approach with the Human Approval Strict (HA-S) technique presented in~\cite{saisubramanian2022avoiding} which avoids all types of NSEs.

\subsubsection{Classifier Training} We use similar strategy as \textit{Markov-HA-S} in training the classifier $\hat{C_i}(s_t, a_t)$, using random query data. Note that for our training data, ${C_i}(s_t, a_t)$ is an indicator function specifying whether NSE of class $i$ is present while \textit{Markov-HA-S} requires it to be a numerical score. For boxpushing domain, our classifier was trained with $\sim$1.7k samples as it was enough to achieve close to 100\% classifier accuracy. We compare our method with the \textit{Markov-HA-S} trained with larger learning budgets, i.e. \{2k, 4k, 6k, 7k\} samples, and we report its average performance over these training budgets. This ensures that variation in a single \textit{Markov-HA-S} learning budget does not unfairly affect our comparisons. For driving domain, our classifier was trained with $\sim$3.5k samples (our classifier achieved close to 99\% accuracy) and compared against the \textit{Markov-HA-S} trained with learning budgets \{4k, 6k, 7k\}.

\subsubsection{Results and Discussion} As we are using Markovian NSEs for the Grid-World experiments, we report the NSE values using the same NSE penalty as \textit{Markov-HA-S}: fixed numerical value 5 and 10 for mild and severe NSEs respectively. We adapt our MBGE method to reduce the Markovian NSEs. For our model-free variant, classifier’s output $\hat{C_i}(s, a)$ is treated as a proxy for true costs $c(s,a)$ and the accumulated cost over the trajectory is to be minimized. The performance (average $\pm$ one standard deviation) of this model-free variant is reported in Table~\ref{table:grid_results} as \textit{MF-Lagrange}, alongside MBGE and \textit{Markov-HA-S}. The reported values are aggregated over five different problem instances for each task.

% we Figure~\ref{fig:grid_expt} compares the performance between our MBGE (Model-Based Gradient Estimation) method and \textit{Markov-HA-S}. The rectangular marker indicates the average value while the whisker represents one standard deviation. The reported values in Figure~\ref{fig:grid_expt} are aggregated over five different problem instances for each task.

\begin{figure}[t]
	\centering
	\includegraphics[scale=0.42]{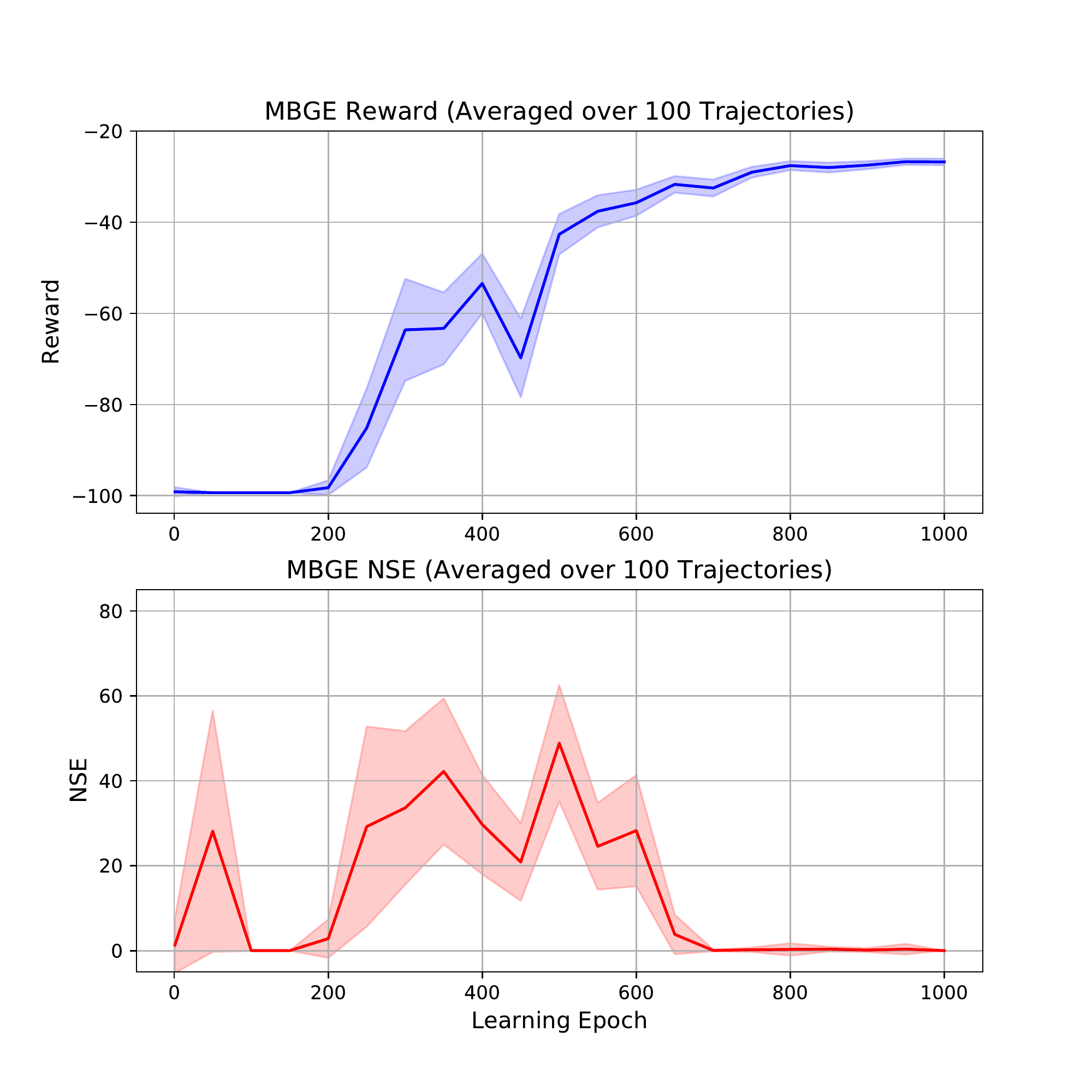}
	\vskip 0pt
	\caption{One MBGE Training Run for Boxpushing}
	\label{fig:grid_mbge}
	\vskip 0pt
\end{figure}

In both tasks, MBGE achieves the highest rewards and lowest NSEs. In boxpushing domain, MBGE's NSE is very close to zero with very low variance. In comparison, \textit{Markov-HA-S} incurs significantly higher NSE with much higher variance and \textit{MF-Lagrange} has serious difficulty in NSE reduction. This agrees with our analysis in Section~\ref{section_mfge},~\ref{section_mbge} since an agent only incurs NSE when it is already pushing the box. \textit{MF-Lagrange} treats classifier output as a blackbox and adjusts all the performed actions in an NSE-inducing trajectory. In contrast, MBGE utilizes the classifier gradient and can fine-adjust the actions only at timesteps when agent is pushing the box, resulting in better NSE performance. In autonomous driving domain, MBGE manages to reduce the average NSE to a very small value with lower variance. These results support our claim where our method is general enough to be applied to Markovian NSEs and yet achieve better performance than baselines.

To better understand how MBGE learns a safe policy, we plot a single MBGE training run in Figure~\ref{fig:grid_mbge}. Every 50 epochs, the policy is executed in test environment and the reported reward and NSE are averaged over 100 test trajectories. The semi-transparent band indicates $\pm 0.5$ standard deviation. In the first 200 epochs, the agent explores the environment and has not discovered that the box can be picked up to achieve higher reward, thus both the reward and NSE are low. After epoch 200, the agent learns to push the box to the goal and starts achieving better reward. At the same time, the amount of NSE increases because agent has not learned to avoid dangerous surfaces and this in turn increases the value of Lagrange Multiplier $\lambda_i$. This makes the effect of $\nabla_{\theta} J_{\hat{c}_i}(\pi_{\theta})$ in~\eqref{theta_grad} larger and causes the agent to adjust its policy based on both  $\nabla_{\theta} J_{{R}}(\pi_{\theta})$ and  $\nabla_{\theta} J_{\hat{c}_i}(\pi_{\theta})$. This can be observed toward the later part of training (after epoch 500), where the agent learns to jointly optimize both reward and NSE, eventually converges to a safe optimal policy. 

\subsubsection{Non-Markovian NSE Experiment} We refer the readers to the supplement for additional experiments demonstrating that MBGE better satisfies non-Markovian NSE constraints than \textit{MF-Lagrange} and \textit{Markov-HA-S} baseline.

\subsection*{Continuous Domains}

We experimented with two distinct tasks in continuous planning domains: Navigation~\cite{Faulwasser2009} and Heating, Ventilation and Air Conditioning (HVAC)~\cite{agarwal2010occupancy}. We provide brief descriptions of these two tasks below and refer the readers to~\citet{rddlgym2020} for complete specifications. 

\subsubsection{Navigation} The agent is required to avoid a deceleration zone and move toward a goal position \akshat{in a 2D continuous grid}. Reward is defined as the Euclidean distance between the agent and goal position at every timestep. On top of the existing specifications, we defined an additional \emph{dirty} zone $(2 \leq x \leq 4.5, 0 \leq y \leq 10)$, sitting between the agent's starting position and goal position. The agent incurs no NSE if it passes through the dirty zone fewer than two timesteps throughout the entire trajectory (mild NSE: $2-3$ timesteps, severe NSE: $\geq 4$ timesteps in the dirty zone). Our target is to be NSE-free 95\% of the time (i.e. $\mathbb{E}_{\tau \sim \pi_{\theta}} [ C_{0}(\tau) ] \geq 0.95$). 

\subsubsection{HVAC} In HVAC control, the agent controls the amount of heated airflow into a set of inter-connected rooms. The original problems specifies room temperature to be between $20^\circ C$ to $23.5^\circ C$. It also incentivizes the agent to maintain the room temperatures at the midpoint of this temperature range (around $21.75^\circ C$). Unbeknownst to the MDP designer, one of the rooms is actually a server room and there is a safety requirement to further keep the temperature of this room below $21^\circ C$. As such, our non-Markovian NSEs are defined as follows: a trajectory is NSE-free if the server room temperature is higher than $21^\circ C$ for no more than one \emph{consecutive} timestep (mild NSE: $2-3$ consecutive timesteps, severe NSE: $\geq 4$ consecutive timesteps above $21^\circ C$). The NSE constraint is to have no-NSE 95\% of the time (i.e. $\mathbb{E}_{\tau \sim \pi_{\theta}} [ C_{0}(\tau) ] \geq 0.95$). 

\begin{figure}[t]
	\centering
	\includegraphics[scale=0.47]{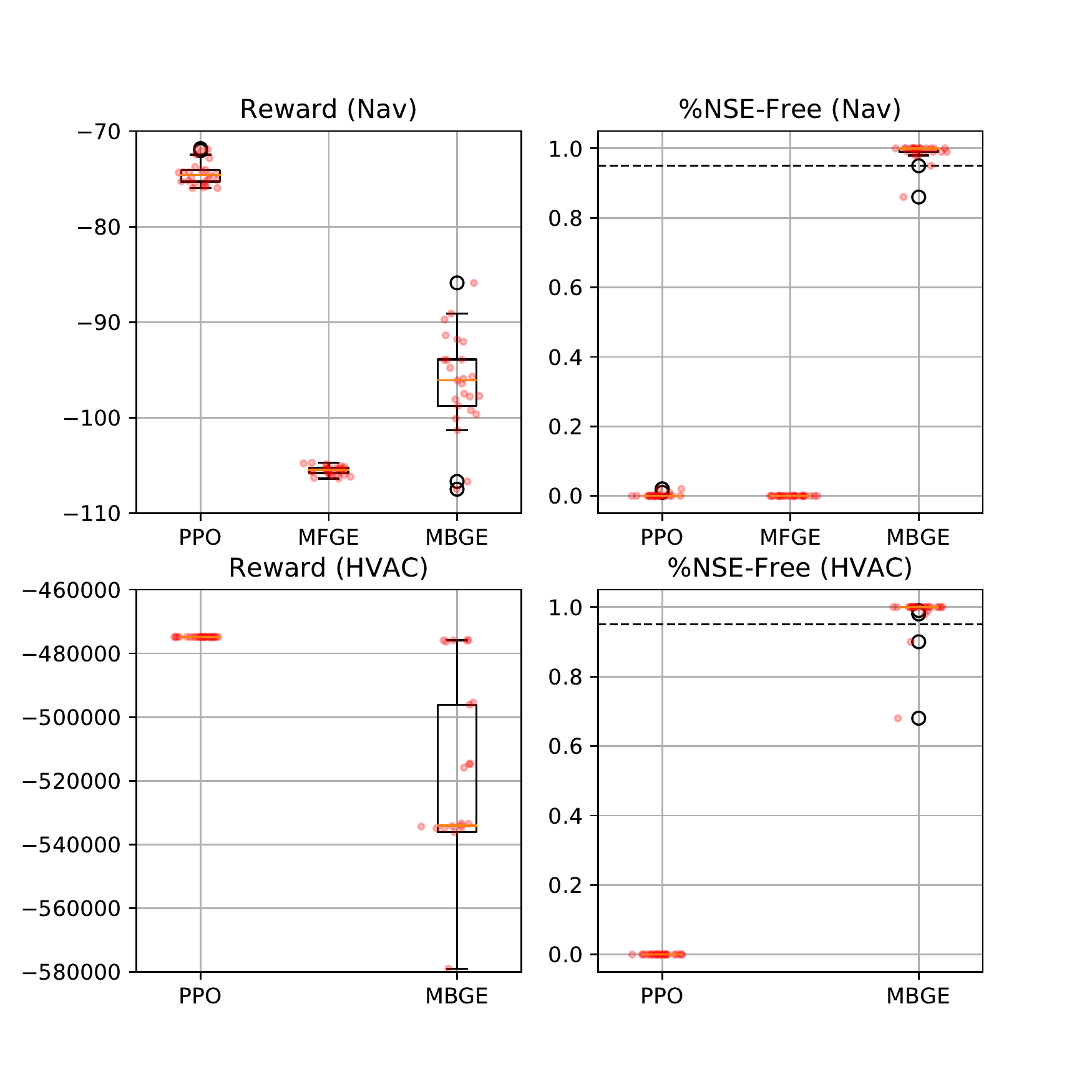}
	\vskip 0pt
	\caption{Continuous Domain Experiments (box plots)}
	\label{fig:rddl_expt}
	\vskip -5pt
\end{figure}

\subsubsection{Classifier Training} The NSEs defined for the two tasks above are non-Markovian NSEs in nature and the whole trajectory needs to be inspected to determine the presence of an NSE, i.e. $C_{i}(\tau)$. As such, we use the GRU-based classifier $\hat{C}_{i}(\tau)$ illustrated in Figure~\ref{fig:SCG} and report the percentages of test trajectories actually having no NSE, i.e. $\mathbb{E}_{\tau \sim \pi_{\theta}} [ C_{0}(\tau) ]$. For both navigation and HVAC tasks, we trained the classifier using close to 200k labelled trajectories, achieving around 99\% classification accuracy.

\subsubsection{Results and Discussion} The boxplots in Figure~\ref{fig:rddl_expt} report the total reward and percentage of NSE-Free Trajectories achieved by policies trained using PPO, Model-Free Gradient Estimation (MFGE) and Model-Based Gradient Estimation methods (with the latter two being the proposed methods in this paper). The charts present the policy performance within their last 200 epochs in five independent training runs. 

From the top half of Figure~\ref{fig:rddl_expt} (for navigation domain), PPO achieves the best reward while incurring NSE almost 100\% of the time. This is expected since PPO simply optimizes reward, without being aware of the presence of NSEs. Our proposed MBGE method is able to strike good balance between reward optimization and NSE avoidance, achieving average reward around -95 while almost always satisfy the NSE constraint. The MFGE method significantly underperforms, collecting lower reward and never meets the NSE constraint. The high-variance nature of MFGE method creates difficulty in practice. Furthermore, the NSE we defined is typically observed over a smaller subpart of the trajectory. As discussed in Section~\ref{section_mfge}, MFGE tends to overcorrect the entire trajectory and cannot perform targeted adjustment. %The episodic MDP task we tested involves many timesteps (e.g. 40 timesteps for HVAC), making it more difficult for MFGE to work.

\begin{figure}[t]
	\centering
	\includegraphics[scale=0.45]{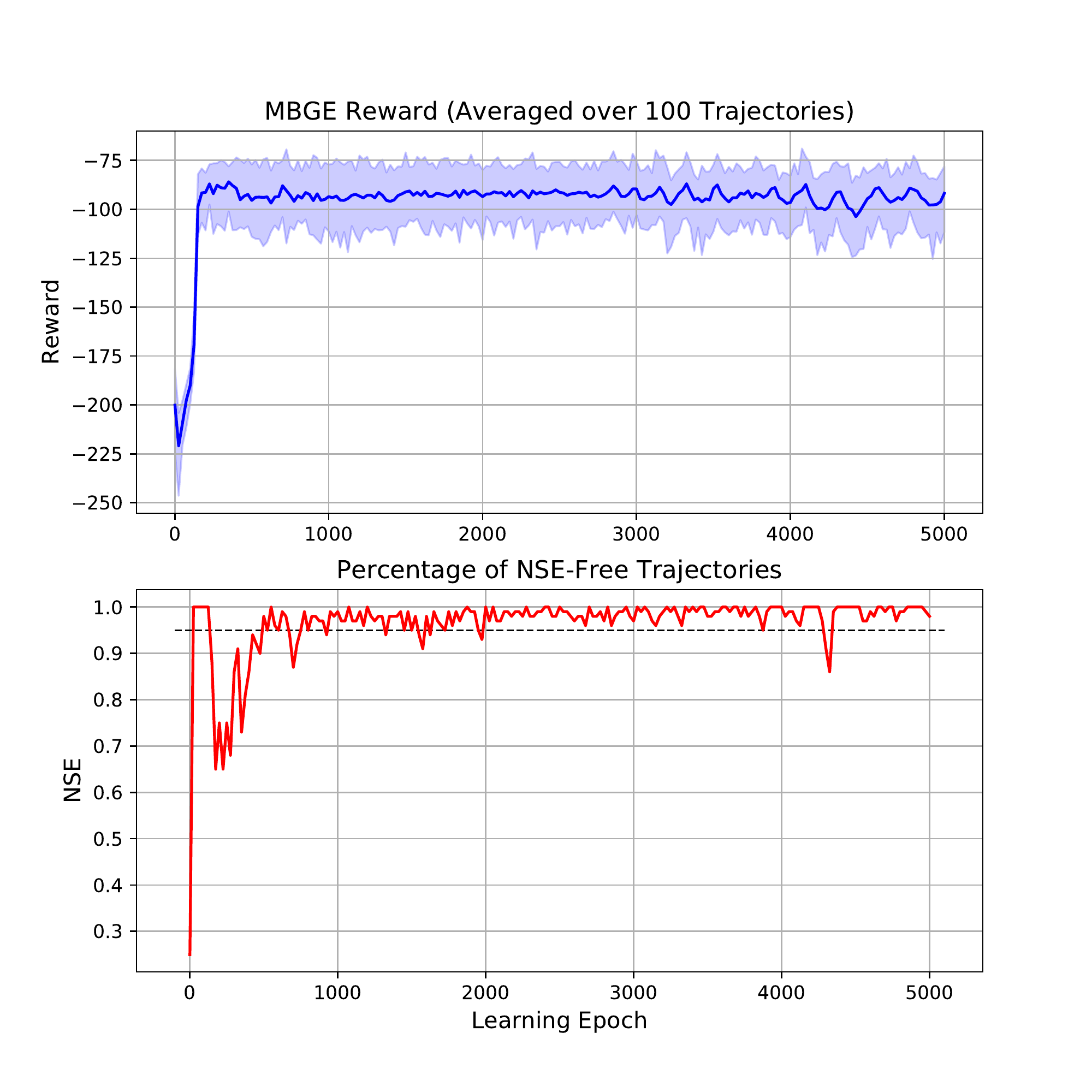}
	\vskip 0pt
	\caption{One MBGE Training Run for Navigation}
	\label{fig:rddl_nav}
	\vskip 0pt
\end{figure}

We do not include MFGE in HVAC performance reporting since it fails to converge to the constrained local optimum in all training runs. The bottom half of Figure~\ref{fig:rddl_expt} shows similar pattern, with PPO being able to maximize reward but violating constraint all the time. For MBGE, most runs converge to safe policy with average returns around -520k, supporting our claims that MBGE enables agent to optimize reward and NSEs jointly.

Figure~\ref{fig:rddl_nav} illustrates the MBGE learning process for the navigation domain (figure for HVAC can be found in the supplement). The solid blue line indicates the total reward averaged over 100 test trajectories and the semi-transparent band outlines the maximum and minimum reward among the 100 test trajectories. As the agent learns the move toward to goal, it starts to incur NSE (around epoch 150) since the shortest path would pass through the dirty zone. This increases $\lambda_i$ and intensifies the effects of NSE term $\nabla_{\theta} J_{\hat{c}_i}(\pi_{\theta})$ during policy update. Armed with the gradient information from the learned classifier, MBGE optimizes total return and NSE jointly and discovers a safe path bypassing the dirty zone (around epoch 500). The red curve demonstrates that the converged policy is NSE-free at least 95\% of the time.

% \begin{figure}[t]
% 	\centering
% 	\includegraphics[scale=0.45]{Figures/rddl_hvac6.pdf}
% 	\vskip 0pt
% 	\caption{One MBGE Training Run for HVAC}
% 	\label{fig:rddl_hvac}
% 	\vskip 0pt
% \end{figure}

\section*{Conclusion}

We have presented a method for safe MDP planning that avoids negative side effects (NSEs), which may arise during policy execution based on an incomplete model of a complex real world environment. Unlike previous works that require knowledge of numerical safety cost functions, our method learns a RNN-based classifier that learns to label state-action trajectories with different safety categories based on collected NSE dataset. Thus, our method can address a rich set of non-Markovian NSEs, unlike previous works which are limited to Markovian safety cost functions. Furthermore, we developed a model-based method that integrates the differentiable classifier with the MDP model to estimate the gradient of the classifier w.r.t. policy parameters, which is a better approach than an entirely model-free way of estimating gradients. Empirically, our method worked significantly better than a number of baselines.

\section*{Acknowledgments}

This research/project is supported by the National Research Foundation Singapore and DSO National Laboratories under the AI Singapore Programme (Award Number: AISG2-RP-2020-016).

%%%%%%%%%%%%%%%%%%%%%%%

%\clearpage
\small
\bibliography{aaai23}
\end{document}

\maketitle
\thispagestyle{empty}

\section{Hyperparameter Settings for Grid World Experiments}

\begin{table}[th]
\begin{center}
\begin{tabular}{ | c | c | c | c | c | c |  }
\hline
\textbf{Neural Network Model} & \textbf{Hidden Units} & \textbf{Normalization} & \textbf{Learning Rate} & \textbf{Minibatch Size} & \textbf{Others} \\
\hline
Classifier & [32, 32] & Batch Norm & 0.0001 & 100 & Dropout: 0.5 \\
Policy Net & [32, 32] & Layer Norm & 0.0003 & 100 & PPO Clip: 0.2 \\
Value Net & [32, 32] & Layer Norm & 0.0003 & 100 & - \\
\hline
\end{tabular}
\caption{Hyperparameter Settings of Neural Network Models for Grid World Experiments}
\label{tableS:hyperparam}
\end{center}
\end{table}

The initial value of Lagrange Multiplier $\nu_i$ used is 1.0. The learning rate for $\nu_i$ is 0.0003 for both boxpushing and driving tasks. 

\section{Hyperparameter Settings for Continuous Domain Experiments}

\begin{table}[th]
\begin{center}
\begin{tabular}{ | c | c | c | c | c | c |  }
\hline
\textbf{Neural Network Model} & \textbf{Hidden Units} & \textbf{Normalization} & \textbf{Learning Rate} & \textbf{Minibatch Size} & \textbf{Others} \\
\hline
GRU Classifier & [64, 64] & - & 0.0001 & 100 & Dropout: 0.5 \\
Policy Net & [64, 64] & Layer Norm & 0.0003 & 100 & PPO Clip: 0.2 \\
Value Net & [64, 64] & Layer Norm & 0.0003 & 100 & - \\
\hline
\end{tabular}
\caption{Hyperparameter Settings of Neural Network Models for Continuous Domain Experiments}
\label{tableS:hyperparam}
\end{center}
\end{table}

The initial value of Lagrange Multiplier $\nu_i$ used is 1.0. The learning rates for $\nu_i$ is 0.003 (Navigation task) and 0.005 (HVAC task).

\section{Additional Experiment Results}

\begin{figure*}[h!]
	\centering
	\includegraphics[scale=0.8]{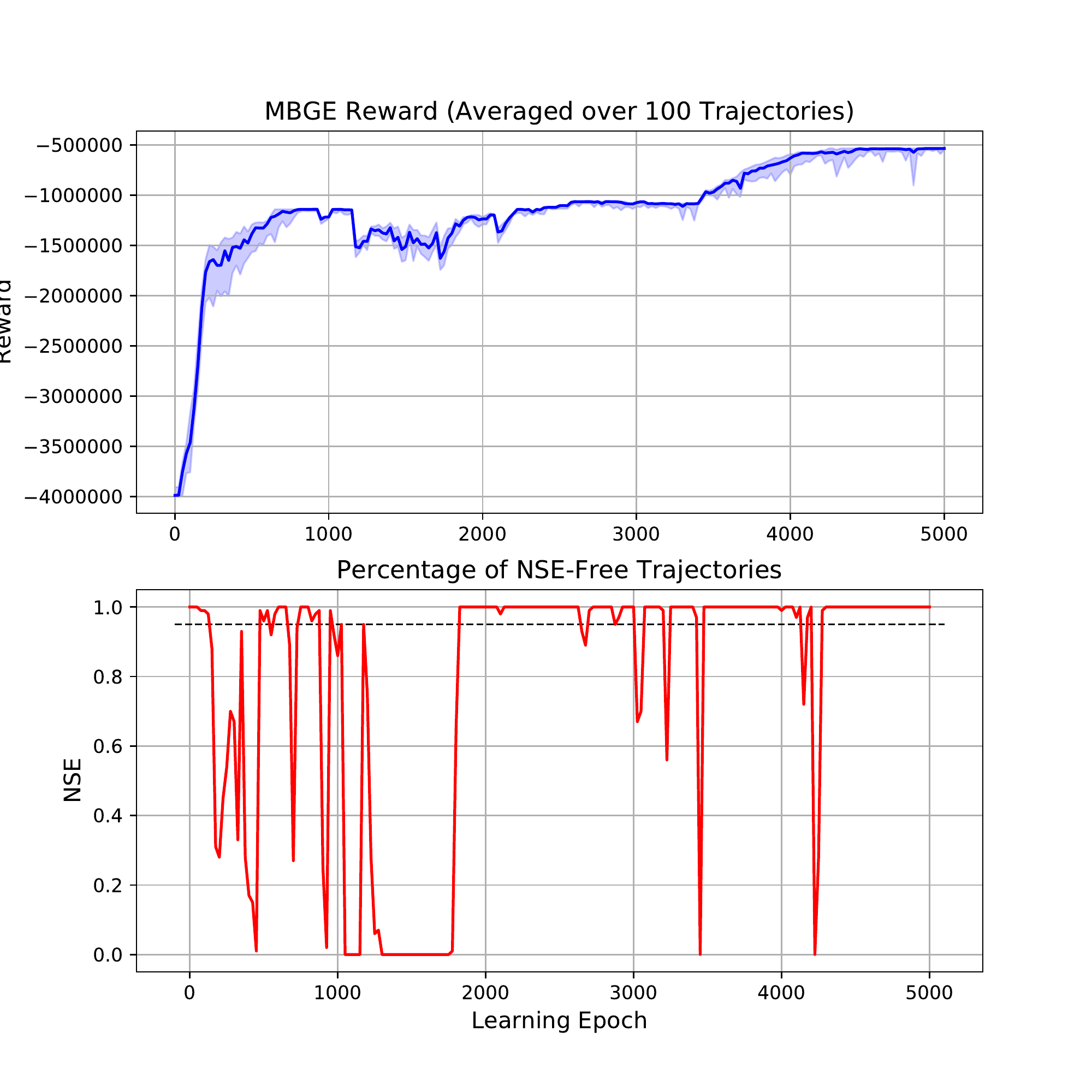}
	\vskip 0pt
	\caption{One MBGE Training Run for HVAC}
	\label{fig:rddl_hvac}
	\vskip 0pt
\end{figure*}

Figure~\ref{fig:rddl_hvac} illustrates the MBGE learning process for the HVAC task in continuous domain experiment. 

\section{Hardware \& Software Specifications}

All experiments were conducted on an AMD EPYC 7371 16-Core CPU machine with 512 GB memory, running Ubnutu 20.04 LTS. The Python package of PyTorch v1.11 was used to train the neural networks.

\section{Code Appendix}

Source code used in conducting the empirical experiments is available on \url{https://github.com/siowmeng/Avert-NonMarkovNSE}.